\documentclass[11pt]{article}

\usepackage{tabularx}
\usepackage{acl2013}
\usepackage{graphicx}
\usepackage{algorithmic}
\usepackage{times}
\usepackage{latexsym}
\usepackage{multirow}
\usepackage{url}
\usepackage{subfigure}
\usepackage{array}
\usepackage{sparklines}
\usepackage{amsmath,amsfonts,amssymb,amsthm}

\usepackage{paralist}
\usepackage{color}

\newcommand{\comment}[1]{}

\newcommand{\etal}{\mbox{\it et al.}}

\newcommand{\eg}{\mbox{\it e.g.}}

\newcommand{\bi}{\begin{list}{$\bullet$}{
    \setlength{\leftmargin}{1.5 em}
    \setlength{\itemsep}{0 pt}
    \setlength{\topsep}{3 pt}
    \setlength{\parsep}{3 pt}
    \setlength{\partopsep}{0 pt}
    \setlength{\labelwidth}{1 em}
    \setlength{\labelsep}{0.5 em}
    \setlength{\parskip}{0cm}  }}
\newcommand{\ei}{\end{list}}

\newcommand{\BE}{\begin{enumerate}}
\newcommand{\EE}{\end{enumerate}}

\newcommand{\initab}{                           
\begin{tabbing}
XXX \= XXXX \= \kill
}
\newcommand{\begpub}{
\begin{quotation}
\noindent
}

\newcommand{\finpub}{
\end{quotation}
}

\def\upcase{\expandafter\makeupcase}

\newcommand{\sys}{\mbox{\sc CrowdAnno}}   

\newcommand{\eat}[1]{}

\title{Visualizing NLP annotations for Crowdsourcing }

\author{Hanchuan Li, Haichen Shen, Shengliang Xu and Congle Zhang\\
 Computer Science \& Engineering \\
 University of Washington\\
 Seattle, WA 98195, USA \\
 {\tt \{hanchuan,haichen,shengliang,clzhang\}@cs.washington.edu} \\}

\begin{document}

\maketitle

\begin{abstract}
Visualizing NLP annotation is useful for the collection of training data for the statistical NLP approaches.  Existing toolkits either provide limited visual aid, or introduce comprehensive operators to realize sophisticated linguistic rules. Workers must be well trained to use them. Their audience thus can hardly be scaled to large amounts of non-expert crowd-sourced workers. In this paper, we present \sys, a visualization toolkit to allow crowd-sourced workers to annotate two general categories of NLP problems: clustering and parsing. Workers can finish the tasks with simplified operators in an interactive interface, and fix errors conveniently. User studies show our toolkit is very friendly to NLP non-experts, and allow them to produce high quality labels for several sophisticated problems. We release our source code and toolkit to spur future research.
\end{abstract}

\section{Introduction}
Statistical machine learning approaches has made great successes in research disciplines such as parsing~\cite{klein2003accurate}, information extractions~\cite{banko2007open}, question answering~\cite{kwok2001scaling}. Significant advances have been made in recent years on inferrring latent labels for sequence observations using neural networks~\cite{socher2011parsing,Huang14}  and kenerl methods~\cite{zhao2012adaptive}. Yet off-the-shelf models learned from training data of one particular domain (usually newswire) would often underperform at present tasks, whose data could come from other domains such as social media and biomedical data~\cite{wu2014learning}. To fully exploit the power of statistical approaches, it is useful to quickly collect plentiful in-domain training data in a short period of time.

With the advent of crowd-sourcing platforms (e.g. Amazon Mechanical Turk\footnote{http://www.mturk.com} and Odesk \footnote{http://www.odesk.com}), it becomes realistic to quickly hire a large group of crowd-sourced workers with a fair amount of cost. For example, there are over 500,000 workers in AMT and in average, they are more educated than the United States population. But when the NLP data is annotated by unevenly-trained non-expert workers, errors are rampant. We believe they happen for at least three reasons: (1) sophisticated linguistic practices, for example, the guidelines for penn treebank are more than 300 pages. It is impossible for non-experts to catch all the details; (2) comprehensive operators caused by sophisticated rules, for example, to generate a parsed tree, annotators must identify the POS of each tree node from a set of hundreds tags; (3) many NLP problems are structured prediction problem, when the labels depend on each other, each decision requires deep lookahead and backtracking.

Therefore, it is not surprising that the outputs of the crowd-sourced workers are far from oracles. During annotations, they are having limited NLP knowledge, are blind about the dependencies of the data, but having excessive amount of options to choose. To alleviate the problem, we believe a good visualization toolkit for crowd-sourced workers could dramatically increase their productivity and accuracy. Unfortunately, existing toolkits either provide limited visual aid, or introduce comprehensive operators to realize sophisticated linguistic practices and restrict their pool of workers. Instead of anticipating the workers to be NLP experts, we aim to allow ordinary people to annotate the training corpus. For this, the toolkit is designed with several principles in mind: (1) simplified operators: comprehensive operators for crowd workers often mean longer education time and lower accuracy. We thus only allow users to {\em click} and {\em draw} over the data. The tradeoff between expressiveness and quantity is worthwhile because we can hire crowd-sourced workers to generate more training data with less errors. (2) interactive interface for deep lookahead and backtracking, workers should efficiently read the dependency of the data in some interactive way; (3) convenient trial and error, for structured prediction, labels are related to each other while human beings have to annotate them in some serialized manner. So it is inevitable that workers would frequently fix their errors; (4) generalization, many toolkits only specialize on one problem because they must provide comprehensive and specialized operators. As oppose to them, our toolkit with simplified operators is consequently easier to cover a broader range of NLP tasks.

In this paper, we present \sys, a visualization toolkit for crowd-sourced NLP annotation. The key observation is that many NLP problems turn out to be clustering and parsing. On one hand, NLP system is often asked to identify the relationship between objects, which could be words (\eg\ paraphrasing), mentions (\eg\ co-reference), sentences (\eg\ summarization), documents (\eg\ sentiment analysis), etc. All these tasks could be treated as some practice of clustering or formulated as a special form of reinforcement learning~\cite{Huang12}. On the other hand, many NLP applications rely on the results of parsing. For example, relation extraction~\cite{hoffmann2011knowledge} approaches use the dependency features generated from the parsed trees; some syntax-based translation method translate between trees of different languages. \sys\ allows crowd-sourced workers to generate the clustering graphs and the parsing trees. They can annotate the training data with simplified operators to avoid tedious education; they can backtrack their labels in an interactive interface to better read the data; they can easily edit their previous actions to quickly fix the errors. User studies show our toolkit is very friendly to NLP non-experts, and allow them to produce high quality labels for several sophisticated problems. 

The rest of this paper is organized as follows. In Section 2 we demonstrate the overview of our toolkit. Section 3 and 4 presents the implementations for clustering and parsing annotation respectively. In Section 5 we introduce the evaluation metrics, and present the experimental results of our toolkit with respect to several baselines. In section 6 we discuss the related work of our toolkit. Finally we conclude our work and present the discussion.

\section{Toolkit Overview}

In this section, we first introduce the design decisions and the
overview of the toolkits. Then we present the user interface and the
major functions.

The toolkit is designed and implemented as a web application, since it
is easiest for crowd workers to visit.

\begin{figure}
\centering
\includegraphics[width=3.1in]{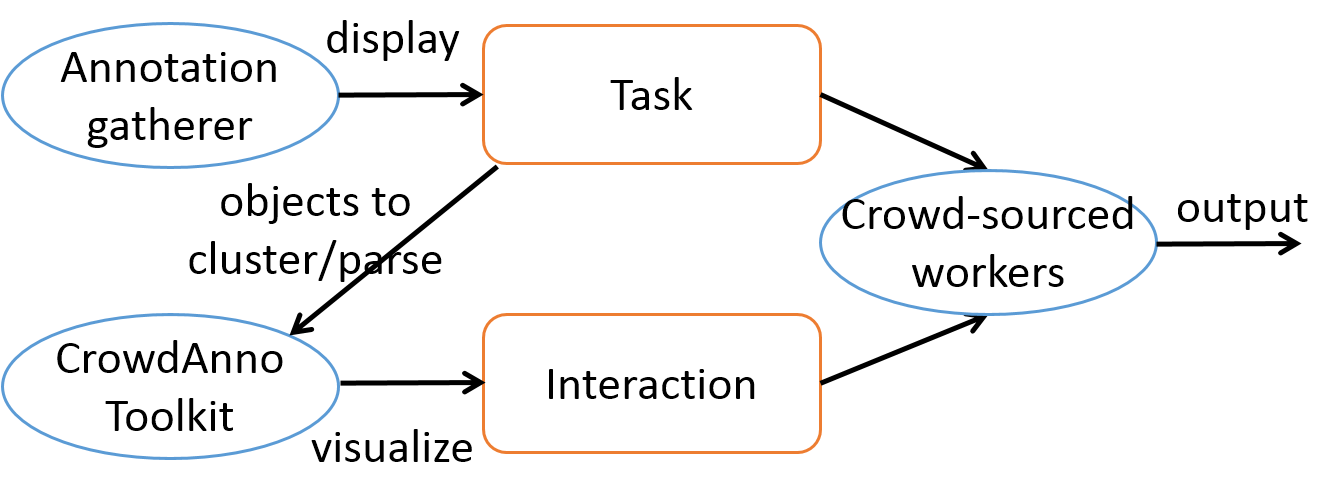}
\caption{Workflow of the toolkit: annotation gatherers generate the
html pages to display the task; our toolkit take the html pages as the
input and visualize the interactions; crowd-sourced workers see the
integrated interface and generate the outputs. }
\label{fig:workflow}
\end{figure}

The users of our visualization toolkit include {\em annotation
gatherers} and {\em crowd-source workers}. In general, annotation
gatherers (NLP experts and well-trained programmers) use our toolkit
to collect training data from crowd-source workers for their NLP
systems. We believe annotation gatherers know their NLP problems
better than anyone else, so we leave them to implement a html page to
display their problem. The html page would call the APIs to pass data
(\eg\ objects to cluster) to the toolkits. The core component then is
to visualize the problem, as well as allowing workers to interact with
the objects. Our toolkit keeps the task-visualization transparent to
maximize the flexibility of our toolkit. Figure~\ref{fig:workflow}
shows the workflow of our toolkit.

\subsection{Cluster Labeling}

\begin{figure*}[th]
\centering
\includegraphics[width=6.1in]{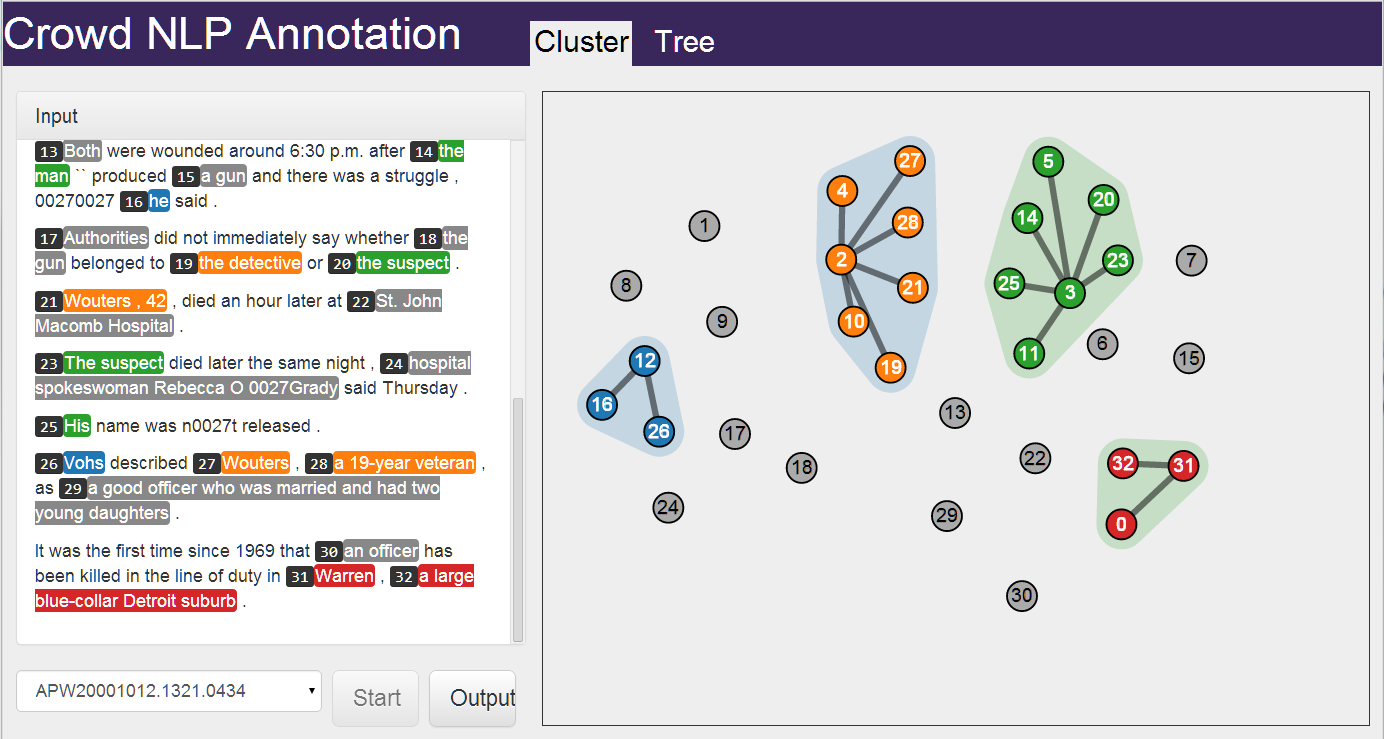}
\caption{Interface of clustering: the left side displays the NLP task, which takes the input from annotator gatherer; the right side is the interaction area, where workers}
\label{fig:interface1.png}
\end{figure*}

We use co-reference as the running example for clustering
visualization. Figure~\ref{fig:interface1.png} shows the user
interface. The left sides (about 40\%) displays the NLP task. One simple way to
display the co-reference task, as shown in the figure, is to highlight
the mentions and to give them unique IDs. The right side of the
interface is for the interaction purpose, where workers generate the
clusters interactively. Each token has a corresponding node in the operation area. 
To bridge the connection between the text display and graphic chart, we label each token and its corresponding 
node with the same index. 

Our main manipulation method for clustering two nodes together is dragging nodes and merging into one group. It is the most intuitive operation for one without any instructions to merge two nodes. We will explain the select and drag, link add and remove, and color scheme in the following paragraphs.

\begin{figure}[tH]
\begin{center}
\includegraphics[height=4cm]{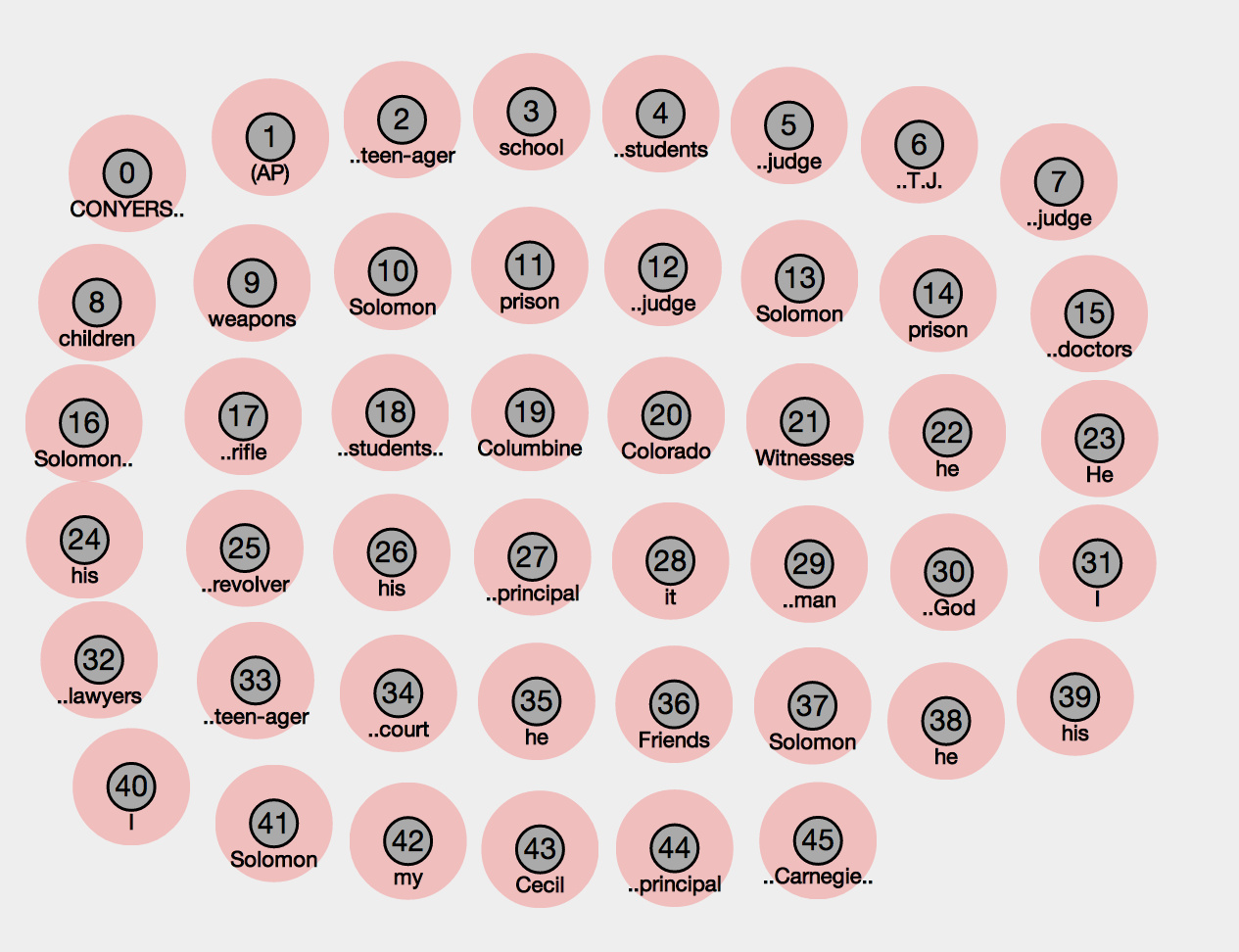}
\caption{Abbreviation text is displayed under each node after dragging starts.}
\label{fig:abbrtext}
\end{center}
\end{figure}

\paragraph{\textsc{Select and Drag}\\}
Since we have two separate parts, one for displaying the text, the other for cluster and group manipulation, a proper selection must be supported in order to reduce user's eye movement between two parts. We adopt three mechanism to address this issue. First, if user click on one token, the corresponding node will be highlighted in red color. Similarly when user starts to drag a node, the token is also highlighted in red background. Second, to accelerate tracing from node to the actual text, once user start to drag a node, the display section will scroll to a proper position to let user easily read the context of this token. Third, as shown in Figure \ref{fig:abbrtext}, when drag event starts, an abbreviation text will be displayed under each node so that user can have a brief concept of what each node represents.

\begin{figure}[tH]
\begin{center}
\includegraphics[height=2.5cm]{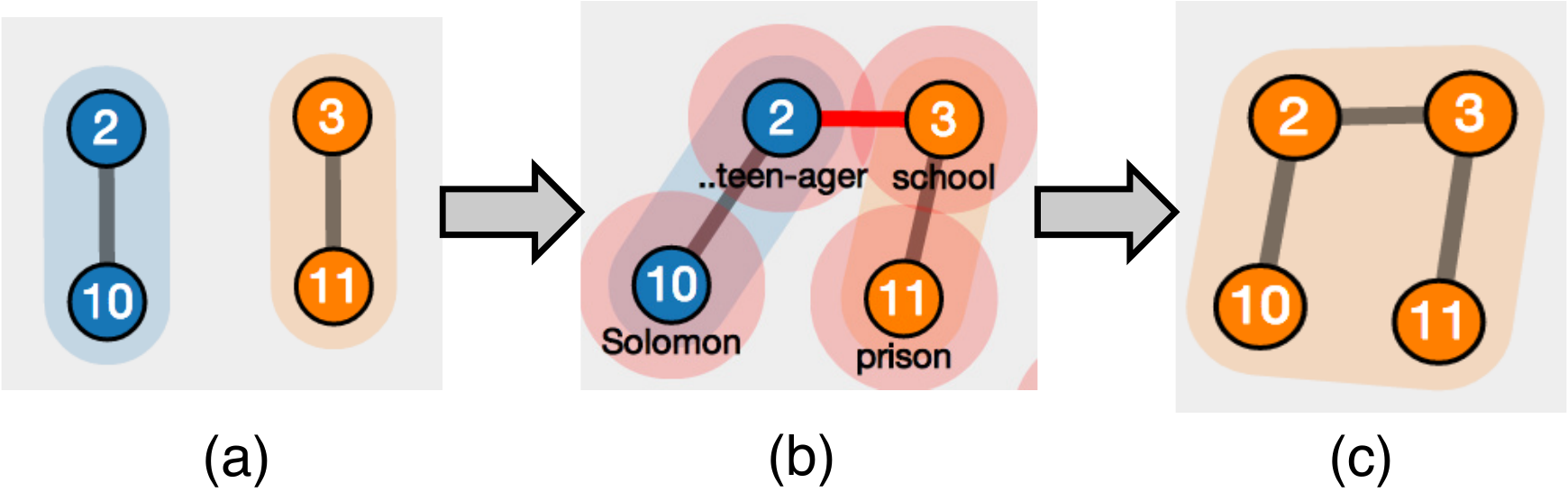}
\caption{(a) before adding the link; (b) in the middle of adding the link; (c) after adding the link.}
\label{fig:addlink}
\end{center}
\end{figure}

\paragraph{\textsc{Link Add and Remove}\\}
Our tool does not allow user explicitly add a link between two nodes. Instead the way to add a link is by dragging a node close enough to the targeted node. Figure~\ref{fig:addlink} shows the process of adding a link. When user starts to drag one node, a shadow circle will be shown around each node. This shadow circle indicates the effective area of adding a link. Once a node is close enough to this shadow circle, a temporary link (in red) will be immediately added to preview the result of this operation (shown in Figure~\ref{fig:addlink}(b)). After user confirms the operation by dropping down the node, a permanent link will be added. Due to the transitivity property of clustering task, \sys\ will automatically bundle two groups together and assign same color for all the nodes in this new group.

\begin{figure}[tH]
\begin{center}
\includegraphics[height=2.5cm]{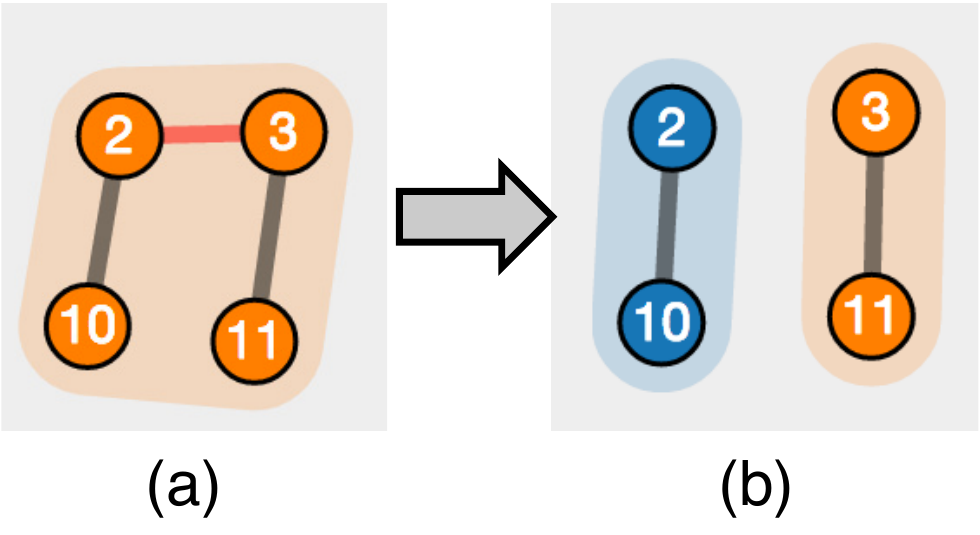}
\caption{(a) in the middle of removing the link; (b) after removing the link.}
\label{fig:rmlink}
\end{center}
\end{figure}

The operation that removes a link is as easy as clicking on the link, shown in Figure~\ref{fig:rmlink}. When the mouse hovers over the link, the link color turns to red to indicate that the link will be removed (Figure~\ref{fig:rmlink}(a)). After the click, link will then be removed. If the group is no longer connected after the removement of that link, the system will find out two connected components and separate the original group into two new groups, e.g., Figure~\ref{fig:rmlink}(b).

\paragraph{\textsc{Color Scheme}\\}
The token and its corresponding node in the graphic sides have the same color. Any change in the color of the node is also reflected in the corresponding token in the text display part (Figure~\ref{fig:interface1.png}). The consistency in color schemes improves the connection between text and graph part.

The default color is grey. If a node (or token) does not belong to any group, we will use the default color as filling color (or background color). We will assign a new color when a group is generated. If two groups merge together, we will keep the color of group being merged to as the color of new merged larger group (Figure~\ref{fig:addlink}). If a group is splited into two, the majority component keeps the color the origin group. 

\paragraph{\textsc{Group Positioning}\\}
To improve the usability and precision of dragging and adding link operations, and to utilize the space effectively, a proper distance should exist between nodes and nodes, and between groups and groups. We use d3 Force model and calibrate the gravity parameter to enforce a proper gap among the nodes. \sys\ also calculate a proper central point for each group, and impose a force to every nodes inside this group towards that direction. Therefore, groups won't be overlapped in a small region.

\subsection{Tree Construction}

The tree annotation task is to construct or edit a tree hierarchy from
a sentence. This task is very common in many sentence based NLP
problems such as POS tagging.

\begin{figure*}
\centering
\includegraphics[width=6.1in]{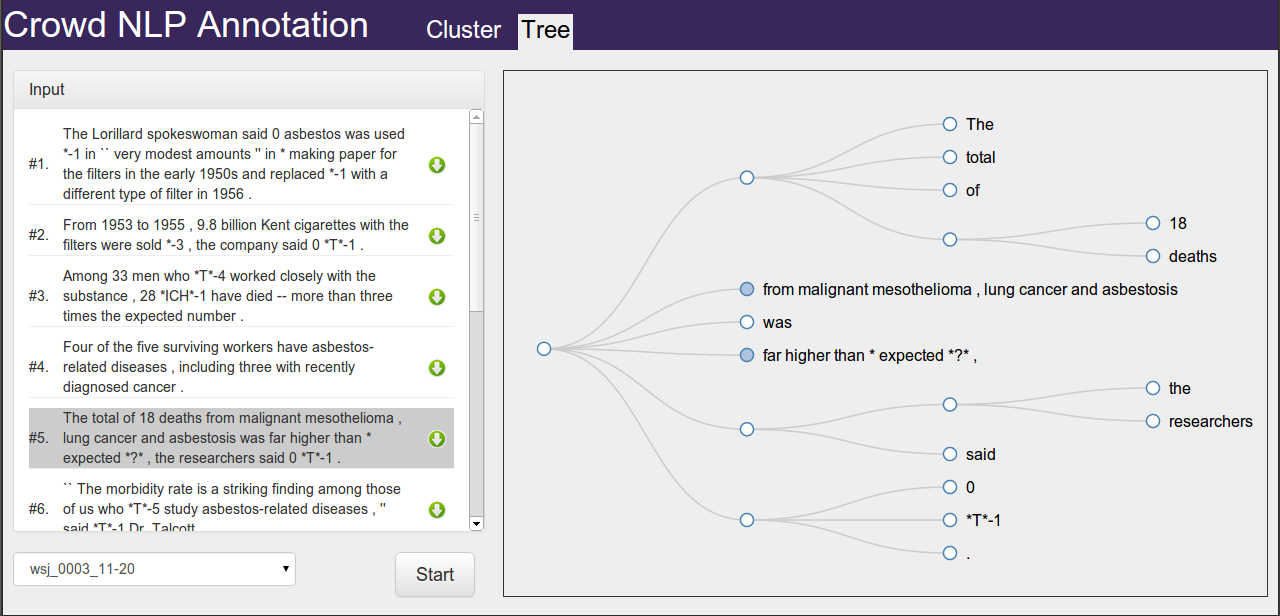}
\caption{The before and after trees in the user interface when parsing
the sentence {``There is no asbestos in our products now."}}
\label{fig:interface_tree.png}
\end{figure*}

\paragraph{\textsc{Interface}\\} The tree annotation task interface
consists of two parts.  We use syntactic parsing as the running
example to show how our toolkit can build a parsed tree.
Figure~\ref{fig:interface_tree.png} shows the general user interface. 

The left part is the sentence input data list. Our tool supports
multiple sentence annotation in the same time. Users can select from a
list of sentence files at the bottom of the left panel. The sentence
files are in plain text format. Each sentence file is simply a list of
sentences, each of which is a line. The words are separated by white
space characters. The sentence list in a given input file will be
shown on the \textit{Input} section. To the right of each sentence,
there is a small \textit{download} sign, on which a click triggers a
download of the current tree built for that sentence.

The right part is the tree editing/construction area. A given input
sentence will be splitted into word nodes vertically. The reason that
we organize the sentence word nodes in this way instead of
horizontally is that each word now always occuppies the same unit
height. It's more space consistent visually and also more efficient in
space usage. Our tool provides three operations for manipulating the
tree construction task.

\paragraph{\textsc{Node addition (grouping)}\\}

\begin{figure}
\centering
\includegraphics[width=1.4in]{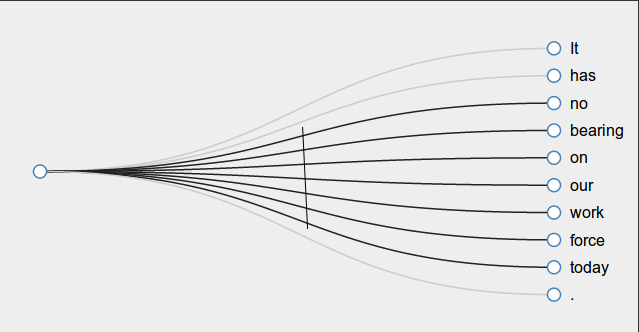}
\includegraphics[width=1.4in]{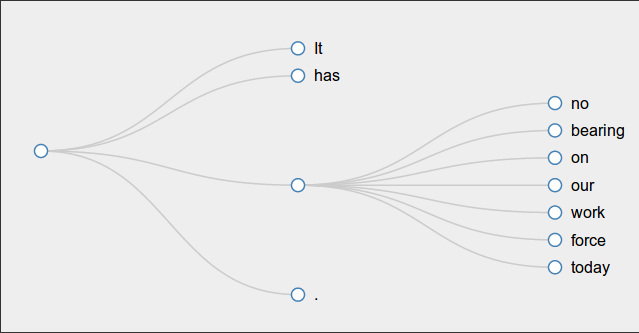}
\caption{Add a new node to group a set of sub-trees by selecting a set
of paths.}
\label{fig:new_node.png}
\end{figure}

Our tool provides a simple {\em line-intersect} operation to group an
existing set of sub-trees as in Figure~\ref{fig:new_node.png}. A new
node which serves as the root of all grouped sub-trees is added as a
result. To the best of our knowledge, we are the first to propose this
visualization operation on tree structures. 

\paragraph{\textsc{Node deletion (ungrouping)}\\}

\begin{figure}
\centering
\includegraphics[width=1.4in]{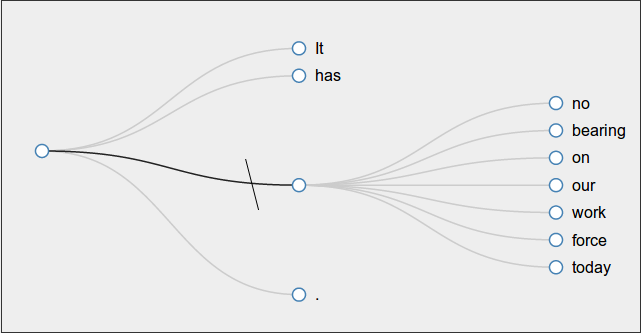}
\includegraphics[width=1.4in]{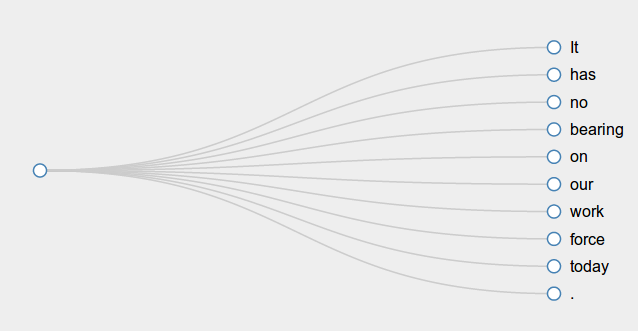}
\caption{Delete a node by selecting a single path.}
\label{fig:node_deletion.png}
\end{figure}

As the couter-operation of node addition, our tool also provides a
node deletion operation also by line intersect. If a single line is
cut, our tool considers it as a node deletion. The node at the child
side of the cut line will be deleted unless the node is a leaf node.
Figure~\ref{fig:node_deletion.png} gives the illustration.

\paragraph{\textsc{Node folding}\\}

\begin{figure}
\centering
\includegraphics[width=1.4in]{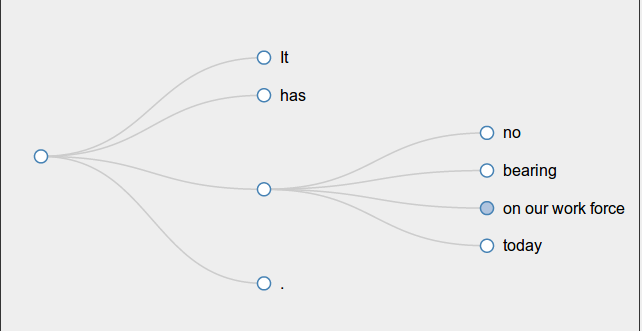}
\includegraphics[width=1.4in]{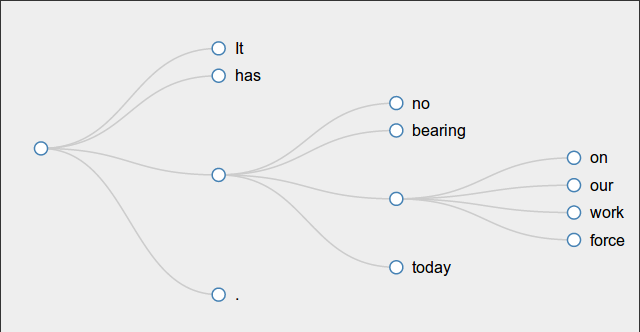}
\caption{Fold a node by clicking on the node.}
\label{fig:folding.png}
\end{figure}

The above two tree editing operations are enough for building arbitray
trees in most NLP tasks. In addition to them, our tool also provides a
node folding operation. It is a change of view operation rather than
an editing operation. Given any non-leaf node, a click on the node
will folds the whole sub tree and represented by a colored leaf node
with the all the words in the sub-tree displayed.
Figure~\ref{fig:folding.png}  gives an illustration.

Since we are targeting sentence based tree builiding tasks, the tree
layout is designed that the node order  is always kept as the input
order, which is the sentence order, after any of the operations. 

%
%
%

\section{Evaluation}
Our goal is to evaluate whether our visualization toolkit is helpful for non-experts of NLP to annotate clustering and parsing training data.

\subsection{Setup}
For clustering, we use co-reference as our running example; for parsing, we ask workers to build syntactic tree for sentences according to their English skills. They don't need to tag any of the nodes. We designed the user studies as follows:
\bi
\item Participants: We asked 6 graduate students aged 21 to 26 years old to participant in our user study. 3 are native English speakers, the other 3 learned English as a second language. None of them are NLP experts. We divide them into native group and foreign group.
\item Instructions: each participant is given two answer examples one on parsing and one on clustering. And given 5 minute to learn the example and ask whatever question they have.
\item Clustering: each of them are given 10 articles for co-reference with 5 using our visualization tool and 5 conducted using Excel in the traditional way (where they will cluster expressions by putting them into the same row). The orders of the articles and the tools are random.
\item Parsing: each of them are given 10 sentences for tree parsing with 5 using our visualization tool and 5 using text editor to put parenthesis in the traditional way. For example {\em (My dog) (also likes) (eating sausage).)}
\ei
The order of the articles, the sentences and the toolkit they use are randomly shuffled.

We compare the time efficiency and accuracy to generate training data with and without the toolkit. For clustering, time efficiency is defined as the seconds worker used for each correct co-reference cluster. Accuracy is computed by purity. Let $N_{ij}$ be the number of mentions in cluster $i$ that belong to entity $j$, and let $N_j=\sum_i N_{ij}$. Then the purity of a cluster is $p_i=\max_j p_{ij}$. The overall purity is $\sum_i \frac{N_i}{N} p_i$. Purity range between 0 (bad) and 1 (good). Table~\ref{t:cluster} shows the average time, purity for 5 workers. It is clear that using our toolkit dramatically improve the efficiency and accuracy. Purity ranges between 0 (bad) and 1 (good). Table~\ref{t:cluster} shows the average time, purity for 5 workers. It is clear that using our toolkit dramatically improve the efficiency and accuracy.

\begin{table*}[bt]
\begin{center}
\begin{tabular}{|c|c|c|c|c|c|c|c|}
 \hline
 & Participant & W1 & W2 & W3 & W4 & W5 & W6 \\\hline
 &time per entity(second) & 7.8 & 6.2 & 6.6 & 9.4 & 10.3 & 11.0\\
Text Editor& purity & 0.82 & 0.76 & 0.88 & 0.61  & 0.78 & 0.82\\\hline
& time per entity(second) & 5.6 & 4.5 & 6.0 & 9.5 & 8.8 & 9.1\\
Visual. Tool& purity & 0.89 & 0.86 & 0.91 & 0.72 & 0.87 & 0.83\\\hline
\end{tabular}
\end{center}
\caption{For clustering, comparing time, purity and rand index with and without the visualization toolkit} \label{t:cluster}
\end{table*}

For parsing, time efficiency is defined as the seconds worker used for each token, to compensate the time cost over long sentences.
\begin{table*}[bt]
\begin{center}
\begin{tabular}{|c|c|c|c|c|c|c|c|}
 \hline
 & Participant & W1 & W2 & W3 & W4 & W5 & W6 \\\hline
 Text Editor & time per word (second) & 7.8 & 5.9  & 6.7 & 4.7 & 6.6 & 8.9\\\hline
Visual. Tool& time per word (second) & 4.3 & 4.8 & 4.4& 3.7 & 5.4 & 6.7\\\hline
\end{tabular}
\end{center}
\caption{For parsing, compare time with and without the visualization toolkit} \label{t:cluster}
\end{table*}

Our clustering tool achieved an average 1.0s/15.2\% decrease in time consumption and at the same time an average 6.6\% increase of purity on clustering result. Our tree parsing tool achieved an average of 1.9s/27.8\% decrease in time consumption for parsing. We manually checked the quality of these tree and find out that the qualities with and without the toolkit is quite comparable. It is reasonable because we provide workers text edits with the function of parenthesis matching, so people can easily figure out the errors of parenthesis in the sentences. It shows that people are more intolerable about the errors in parsing, even with text editors. But it is clear that they spent much more time on each tree. During crowd-sourcing, time cost equals money spending. It shows that our toolkit is still valuable.

\section{Related Work}

Many NLP tasks require large amount of high quality training data.
Manual annotation for such training data is well-known for its tedium.
To generate a comprehensive annotated training set requires much human
effort. Annotators are also prone to make mistakes during the long and
tedious annotating process.  Researchers are trying to address these
problems by two means: 1) building specialized annotating tools to
ease the annotating process in the hope of improving efficiency as
well as reducing the error rates; 2) adopting crowdsourcing to scale
up annotating.

\textbf{Specialized annotating tools}. Facing one of the biggest common problems, many NLP researchers have developed a number of tools
for annotating training corpora along the history of NLP research. At
  first, before the blossom of the web, tools are generally built as
  local programs such as the WordFreak linguistic annotation
  tool~\cite{Morton2003WOT} and the  UAM CorpusTool for
  text and image annotation~\cite{ODonnell2008DUC}.
  These tools are very restricted because they cannot scale. Web-based
  annotation tools are developed later in order to scale up the
  annotating process for specific domains, such as anaphora~\cite{Stuhrenberg2007WAA} and  financial reports~\cite{Nan15}. However these tools typically only use very basic HTML based
  techniques to provide very limited visual aids for the annotating
  process. Most related in scope is~\cite{yan2012collaborative} which
  provides a collaborative tool to assist annotators in tagging of
  complex Chinese and multilingual linguistic data. It visualizes a
  tree model that represents the complex relations across different
  linguistic elements to reduce the learning curve. Besides it
  proposes a web-based collaborative annotation approach to meet the
  large amount of data.  Their tool only focuses on a specific area
  that is complex multilingual linguistic data, whereas our work is
  trying to address how to generate a visualization model for general
  data sets.

\textbf{Crowdsoursing in NLP}. Crowdsourcing \cite{howe2006rise} is a
popular and fast growing research area. There have been a lot of
studies on understanging what it is and what it can do. For instance,
\cite{quinn2009taxonomy} categorizes crowdsourcing into seven genres:
Mechanized Labor, Game with a Purpose (GWAP), Widom of Crowds,
Crowdsourcing, Dual-Purpose Work, Grand Serarch, Human-based Genetic
Algorithms and Knowledge Collection from Volunteer Contributors. Other
works, such as \cite{abekawa2010community} and \cite{irvine2010using},
develops a specific tool and verifies the feasibility and benefit of
crowdsourcing. It is generally convinced that crowdsourcing is of
great beneficial if the tasks are easy to conduct by the workers and
the tasks are independent.

Because of the high labor requirements in typical NLP training tasks,
there also have been some work considering using crowdsoursing in many
NLP tasks. For example, Grady \etal\ generated a data set on document
relevance to search queries for information
retrieval~\cite{Grady2010CDR18666961866723}; Negri \etal\ built a
cross-lingual textual corpora~\cite{Negri2011DCC21454322145510};
Finin \etal\ collected simple named entity annotations using Amazon MTurk
and Crowd-Flower~\cite{Finin2010ANE18666961866709}. Also there are
some researchers observed the hardness of collecting high quality data
and did some studies on improving that, such
as~\cite{Hsueh2009DQC15641311564137}( how annotations should be
selected to maximize quality), and \cite{lease2011quality} (quality
control in crowdsoursing by machine learning).

Different from previous studies, we seek to improve crowdsoursing
annotating quality by greatly lower the usability barrier through the
proposed visualized toolkit rather than trying to cleaning up the data
generated by the crowdsoursing process.

\section{Conclusion and Future Work}
In this paper, we present \sys, a toolkit for crowd-sourced NLP annotation. We visualize two important categories of NLP problems: clustering and parsing. By providing simplified operators and interactive interface, we allow crowd-sourced workers to generate high quality training data for NLP problem. In our evaluation, we let non-expert student to test our toolkit. The results are very promising. Because of the time limitation, we have not yet let the crowd sourced workers to really test our toolkit. In future, we would deploy the toolkit on AMT to collect real training data for NLP problems.

%
\bibliographystyle{naaclhlt2013}
\bibliography{general,paper}

\end{document}